\pdfoutput=1

\documentclass[11pt]{article}

\usepackage[preprint]{acl}

\usepackage{times}
\usepackage{latexsym}

\usepackage[T1]{fontenc}

\usepackage[utf8]{inputenc}

\usepackage{microtype}

\usepackage{inconsolata}

\usepackage{graphicx}

\usepackage{multirow}
\usepackage{enumitem}
\usepackage{booktabs}
\usepackage{adjustbox}
\usepackage{amsmath}
\usepackage{booktabs}
\usepackage{array}
\usepackage{longtable}
\usepackage{xspace}

\usepackage{listings}
\usepackage{xcolor}
\newcommand{\camerareadytext}[1]{}

\newcommand{\fref}[1]{Figure~\ref{#1}}
\newcommand{\tref}[1]{Table~\ref{#1}}

\title{Skill-Aware Data Selection and Fine-Tuning \\
for Data-Efficient Reasoning Distillation}

\author{
Lechen Zhang ~ Yunxiang Zhang ~ Wei Hu ~ Lu Wang \\
University of Michigan, Ann Arbor \\
{ \tt \{leczhang, yunxiang, vvh, wangluxy\}@umich.edu}
}

\begin{document}
\maketitle
\begin{abstract}
Large reasoning models such as DeepSeek-R1 and their distilled variants achieve strong performance on complex reasoning tasks. Yet, distilling these models often demands large-scale data for supervised fine-tuning (SFT), motivating the pursuit of data-efficient training methods.
To address this, we propose a \textit{skill-centric} distillation framework that efficiently transfers reasoning ability to weaker models with two components: (1) \textbf{Skill-based data selection}, which prioritizes examples targeting the student model’s weaker skills, and (2) \textbf{Skill-aware fine-tuning}, which encourages explicit skill decomposition during problem solving. With only 1,000 training examples selected from a 100K teacher-generated corpus, our method surpasses random SFT baselines by +1.6\% on Qwen3-4B and +1.4\% on Qwen3-8B across five mathematical reasoning benchmarks. Further analysis confirms that these gains concentrate on skills emphasized during training, highlighting the effectiveness of skill-centric training for efficient reasoning distillation.

\end{abstract}

\section{Introduction}

Large reasoning models such as DeepSeek-R1~\citep{r1} achieve impressive performance on complex reasoning tasks, yet their\camerareadytext{size and computational} costs remain substantial\camerareadytext{limit practical use}. Distilling these capabilities into weaker Large Language Models (LLMs) via supervised fine-tuning (SFT) is a promising way to broaden the access\camerareadytext{, especially as recent work shows that high-quality small dataset can outperform much larger ones~\citep{ye2025limo}}. A key challenge, however, is the strategy of choosing the right SFT data. Current pipelines typically\camerareadytext{ adopt a one-size-fits-all approach, treating} treat all training examples equally~\citep{r1}, which overlooks the latent structure of data such as the underlying skills\camerareadytext{ and difficulty of examples}, as well as the model’s current knowledge state. In contrast, human learning is highly structured, with knowledge organized hierarchically~\citep{acquisitionofknowledge, researchintolearninghierarchies}. Motivated by this, we ask whether data selection informed by structured relationships among training examples can similarly improve the learning efficiency of LLMs\camerareadytext{ and generalization} in reasoning.

Prior work has made various efforts on structured training, notably by formalizing the notion of LLM \textit{skills}, typically defined as “atomic” competencies (e.g., addition or multiplication) when solving problems~\citep{chen2023skillit, li2025mass}. Some studies~\citep{li2025mass} have explored skill-oriented distillation for LLMs, often by emphasizing broader data coverage\camerareadytext{and quality}~\citep{ye2025limo}. Yet these approaches have two limitations. First, these approaches usually treat each problem as a single unit~\citep{zeng2025evaltree}, ignoring the fact that a single question often involves many atomic skills. Second, prior work~\citep{s1, ye2025limo} generally does not adapt to the model’s current strengths and weaknesses. A geometry-proficient model, for example, learns little from redundant geometry data. Our approach rests on a simple principle: \textbf{LLMs should be trained more\camerareadytext{ intensively} on the atomic skills they struggle with, and less on the ones they already master.}\camerareadytext{ Explicitly identifying and tracking latent skills can enable more targeted training and diagnosis.}

Another challenge lies in enabling LLMs to grasp the hierarchical structure of skills. Conventional distillation exposes models only to QA pairs, leaving relationships among underlying skills implicit. Prior work~\citep{didolkar2024metacognitive} has shown that prompting models with an explicit list of skills can bring improvement\camerareadytext{significantly improve performance}. Inspired by this, we inject structured skill information into training data so that models learn to both solve problems and internalize how different levels of skills are organized.

\begin{figure*}[tb!]
    \centering
    \includegraphics[width=\textwidth]{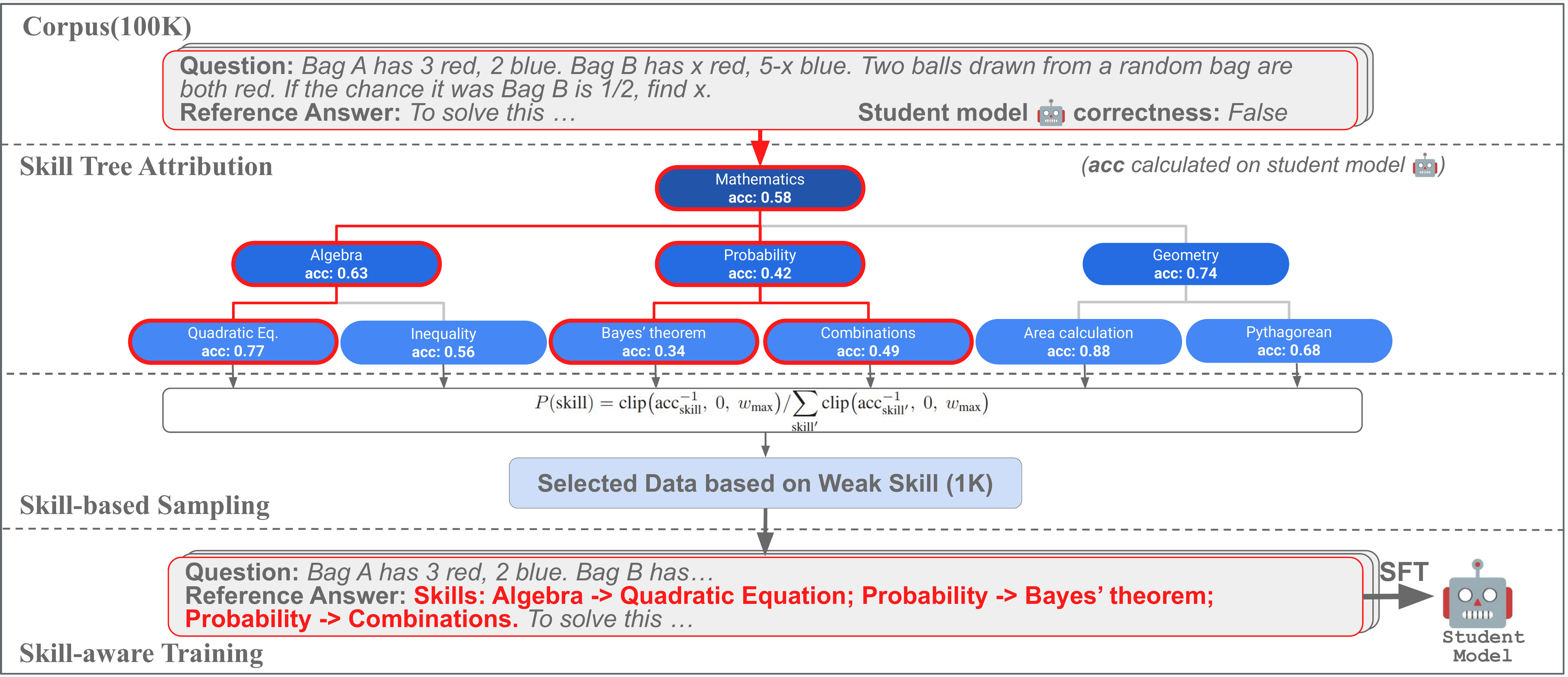}
        \caption{Overview of our skill-centric distillation framework. (1) \textbf{Skill Tree Attribution:} Each problem is mapped to nodes on a hierarchical skill tree~\citep{kaur2025instructskillmix} via top-down LLM-based skill attribution (2) \textbf{Skill-based Sampling:} The student model’s per-skill accuracy guides sampling, with weaker skills emphasized. (3) \textbf{Skill-aware Training:} Selected examples are augmented with explicit skill chains (as shown in red) for skill-aware training.}
    \label{fig:workflow}
    \vspace{-4pt}
\end{figure*}

In this work, we introduce a \textit{skill-centric} data construction framework for math reasoning distillation that leverages a hierarchical skill tree~\citep{kaur2025instructskillmix} to efficiently map examples to multiple relevant skill chains and select targeted data\camerareadytext{ subset} based on per-skill proficiency. Moreover, by embedding interpretable skill chains into the data, the model learns to reason explicitly over a set of skills before answering. Using only 1,000 distillation examples from a 100K corpus, our approach improves Avg@8 accuracy by +1.6\% on Qwen3-4B and +1.4\% on Qwen3-8B across five math reasoning benchmarks, highlighting the promise of structured, skill-centric distillation for LLM reasoning. The code and dataset are available at \url{https://github.com/orange0629/skill-data-selection}.

\section{Related Work}

\paragraph{Distillation of LLM Reasoning}
Knowledge distillation was first introduced as a way to compress large neural networks into smaller ones \camerareadytext{through softened output distributions}~\citep{hinton2015distillingknowledgeneuralnetwork}, later popularized in NLP\camerareadytext{ through models like DistilBERT}~\citep{sanh2020distilbertdistilledversionbert}. Building on these foundations, recent work has shifted toward distilling reasoning abilities. For example, \citet{r1} showed that large models can successfully transfer complex reasoning traces into smaller students\camerareadytext{at scale} with \textasciitilde800K R1 outputs\camerareadytext{, establishing a practical recipe for democratizing reasoning power}. Follow-up studies~\citep{omr, yang2025tabler1inferencetimescalingtable, sun2025tinyr132bpreviewboostingaccuracybranchmerge}\camerareadytext{ like OpenMath-Nemotron~\citep{omr}, Table-R1~\citep{yang2025tabler1inferencetimescalingtable}, Tiny-R1~\citep{sun2025tinyr132bpreviewboostingaccuracybranchmerge}} demonstrate that distilling reasoning\camerareadytext{—whether for proofs, scientific reasoning, or general multi-hop tasks—} can yield compact models that approach\camerareadytext{ or even rival} much larger systems in reasoning capability. However, these works overlook the impact of data selection in distillation, which we address through skill-aware, model-adaptive training.

\paragraph{Data Selection for Efficient Training}
Selecting the most informative training examples has long been studied\camerareadytext{ as a way} to improve model performance under limited\camerareadytext{ data} budgets. Recent studies show that small but carefully curated datasets can yield strong reasoning performance. For example, LIMO~\citep{ye2025limo}, s1~\citep{s1}, and NaturalThoughts~\citep{li2025naturalthoughts} all demonstrate that high-quality and diverse examples often outperform large-scale random sampling. Nevertheless, existing methods largely ignore structured relations among examples, whereas we leverage a skill hierarchy for fine-grained, interpretable selection. \camerareadytext{Building on this insight, structured selection methods~\citep{li2025mass, chen2023skillit}\camerareadytext{such as MASS~\citep{li2025mass} and Skill-It~\citep{chen2023skillit}} leverage graphs and learning dependencies to guide sampling, enhancing efficiency by prioritizing the most instructive reasoning examples. Nevertheless, existing selection methods are not tailored to long-form reasoning distillation where extended reasoning traces demand more targeted supervision.}

\paragraph{Skill Decomposition and Structured Reasoning}
A growing line of work views complex reasoning as a composition of simpler skills and leverages this structure for improved evaluation and training. \citet{didolkar2024metacognitive} showed that prompting LLMs to identify relevant skills improves math performance. EvalTree~\citep{zeng2025evaltree} organizes tasks into a hierarchical skill tree to locate weak skills for synthesizing targeted data. Instruct-SkillMix~\citep{kaur2025instructskillmix} combines pre-defined skills to create\camerareadytext{ diverse} instruction data\camerareadytext{, enabling an 8B model\camerareadytext{ trained on 4K examples} to match far larger ones}. Rather than synthesizing data with skills, our method integrates skills directly into adaptive data selection.

\begin{table*}[]
    \centering
    \resizebox{\textwidth}{!}{%
        \begin{tabular}{l l l ccccc c}
        \toprule
        \textbf{Base Model} & \textbf{Data Selection} & \textbf{Fine-tuning Strategy} & \textbf{AMC23} & \textbf{AIME2024} & \textbf{AIME2025} & \textbf{MATH L5} & \textbf{OlympiadBench} & \textbf{Average} \\
        \midrule
        \multirow{5}{*}{Qwen3-4B} 
         & - & Base & 90.1 & 61.1 & \underline{50.7} & 84.3 & 49.1 & 67.1 \\
         & Full (100K) & Standard SFT & 81.9 & 46.7 & 34.6 & 80.2 & 47.0 & 58.1 \\
         & Random & Standard SFT & 89.5 & 60.1 & 50.3 & 85.3 & 49.0 & 66.8 \\
         & Random & Skill-aware SFT  & \underline{90.9} & 62.2 & 49.9 & \textbf{85.8} & 49.0 & \underline{67.6} \\
         & Skill-based & Standard SFT & 89.1 & \underline{62.5} & 50.0 & \underline{85.5} & \underline{49.5} & 67.3 \\
         & Skill-based & Skill-aware SFT & \textbf{91.9} & \textbf{64.6} & \textbf{50.8} & 85.3 & \textbf{49.6} & \textbf{68.4} \\
        \midrule
        \multirow{5}{*}{Qwen3-8B} 
         & - & Base & 88.2 & 61.1 & 50.2 & 84.7 & 49.1 & 66.7 \\
         & Full (100K) & Standard SFT & 82.4 & 47.1 & 35.5 & 80.6 & 46.7 & 58.5 \\
         & Random & Standard SFT & 90.2 & 62.6 & 50.8 & 86.0 & \underline{50.7} & 68.1 \\
         & Random & Skill-aware SFT  & 91.5 & \underline{65.7} & \textbf{52.6} & \textbf{86.6} & 50.4 & \underline{69.4} \\
         & Skill-based & Standard SFT & \textbf{93.4} & 62.1 & \underline{51.3} & \underline{86.2} & 49.7 & 68.5 \\
         & Skill-based & Skill-aware SFT & \underline{91.9} & \textbf{67.1} & 50.0 & \textbf{86.6} & \textbf{51.6} & \textbf{69.5} \\
        \bottomrule
        \end{tabular}
}
\caption{Accuracy (\%) of Qwen3-4B and Qwen3-8B under different training data selection and fine-tuning strategies using \textbf{1K training examples}. Each column within each base model block is \textbf{bolded} at its highest value and \underline{underlined} at its second highest. Results are reported using Avg@8 across five math benchmarks. Notably, fine-tuning on the full 100K corpus underperforms the base model, highlighting the importance of data selection.%
}
\label{tab:main-results}
\vspace{-4pt}
\end{table*}

\section{Method}

Our approach is motivated by two\camerareadytext{simple} intuitions: (1) models should receive more training data on skills they are weak at, and (2) models\camerareadytext{can} generalize more effectively if they are explicitly trained to recognize\camerareadytext{explicit} skill structures. Our workflow, as shown in \fref{fig:workflow}, begins with a\camerareadytext{curated} corpus of 100K math QA pairs, and a pre-defined skill tree that categorizes mathematical problems into hierarchical skills.

\paragraph{Step 1: Skill tree attribution}
Each training problem is mapped onto the tree by attributing its reference solution to relevant skills. Starting from the root, we prompt \textit{Qwen2.5-32B-Instruct}~\citep{qwen2.5} to decide which high-level skill is involved (prompt shown in Appendix~\ref{sec:tree_attribution_details}). For each selected skill, the LLM is further asked to drill down the decision at the next level\camerareadytext{ of the tree}, until the leaf node is reached. This recursive process leverages the hierarchical structure (with $O(\log N)$ complexity) to avoid overwhelming the model with a flat multi-label decision  and ensures comprehensive coverage of all required skills.\footnote{We manually inspected $\sim$100 random QA pairs and found no evidence of missing or mislabeled skills.}

\paragraph{Step 2: Skill-based sampling}
To adapt training data to a model’s weaknesses, we evaluate the student model on the 100K corpus. For each leaf skill, we compute the model’s accuracy, yielding a skill-wise performance profile. Training examples are then sampled with probabilities inversely proportional to these accuracies:
$
P(\text{skill}) = 
\frac{\text{clip}\!\left(\text{acc}_{\text{skill}}^{-1}, \; 0, \; w_{\text{max}}\right)}
{\sum_{\text{skill}'} \text{clip}\!\left(\text{acc}_{\text{skill}'}^{-1}, \; 0, \; w_{\text{max}}\right)}
$
, where $w_{\text{max}}$ is empirically set to 10,000 to cap divide-by-zero issue. This\camerareadytext{ mechanism} ensures that underrepresented or difficult skills are emphasized while preventing excessive redundancy in well-mastered ones. Using this distribution, we construct our training subsets of 1K.

\paragraph{Step 3: Skill-aware training} Finally, we prepare skill-aware variants of the training data by embedding the explicit skill chain into each instance. For each problem, the ordered sequence of required skills, e.g., \textit{“Skills: [Mathematics → Probability → Bayes' theorem]”}, is prepended before the solution. This encourages the model to explicitly traverse the required skills before attempting the solution,\camerareadytext{ rather than relying on shortcuts,} enabling fine-grained diagnostics of model performance at the skill level.

\section{Experiments}

We conduct a series of experiments to evaluate the effectiveness of our skill-centric pipeline.

\subsection{Setup}
We experiment with two reasoning models\camerareadytext{ from Qwen3 family}: \textbf{Qwen3-4B} and \textbf{Qwen3-8B}~\citep{QwenTeam2025Qwen3}. We extract a clean set of 100K unique QA pairs from \textbf{OpenMathReasoning}~\citep{omr} as our teacher data pool (Details in Appendix~\ref{sec:data_filtering}).
In our experiments, we adopt the existing skill tree hierarchy proposed in the \textit{Instruct-SkillMix} paper~\citep{kaur2025instructskillmix} to label skills for all data.

All models are fine-tuned for 5 epochs\camerareadytext{ unless otherwise noted} (details in Appendix~\ref{sec:training-details}). Evaluation is conducted on five diverse competition-style math benchmarks: \textbf{AMC23}, \textbf{AIME2024}, \textbf{AIME2025}, \textbf{MATH L5} (Level 5)~\citep{math2021}, and \textbf{OlympiadBench}~\citep{olympiadbench}. Avg@8 accuracy (calculated by the average accuracy over 8 independent samples per question) is reported\camerareadytext{; however, for random selection, we averaged the Avg@8 over three random seeds}.

\subsection{Main Results and Analysis}
Table \ref{tab:main-results} shows the performance across different training strategies. We observe that: \textbf{Skill-tree-based data selection generally outperforms random sampling.} For Qwen3-4B, Skill-based data selection yields a +0.5 gain in average accuracy, with the largest improvements on AIME2024 (+2.4). Similarly, Qwen3-8B has a +0.4 average gain with significant improvement on AMC23 (+3.2). These results indicate that aligning training with the model’s weaker skills provides consistent benefits to LLMs.
Second, \textbf{skill-aware training consistently provides additional gains over standard SFT.} Adding explicit skill chains improves average accuracy in nearly all settings, with the largest boost on AIME2024 (up to +5.0), and strongest overall gains of up to +0.8 for Qwen3-4B and +1.3 for Qwen3-8B. Combining Skill-based data selection with skill-aware augmentation further amplifies the effect, yielding significant improvements over random selection (+1.6 for Qwen3-4B and +1.4 for Qwen3-8B) and delivering the strongest overall results\camerareadytext{, including challenging benchmarks such as AIME2024 and OlympiadBench}. These findings confirm that skill-aware sampling and training are complementary and robust. 
Notably, fine-tuning on the full 100K corpus consistently degrades performance relative to the base model, a phenomenon also observed in recent work~\citep{ye2025limo}, highlighting that data selection quality matters more than quantity for effective reasoning distillation.

\begin{figure}[tb!]
    \centering
    \includegraphics[width=\columnwidth]{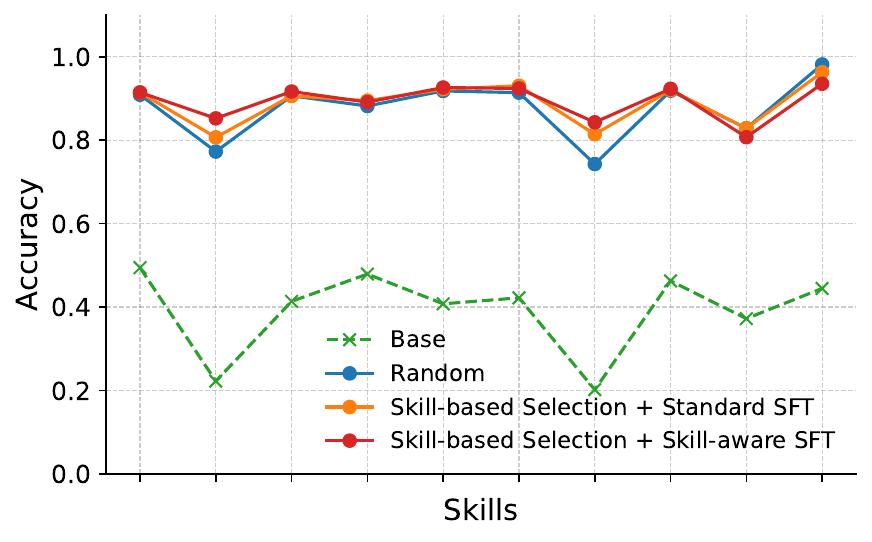}
        \caption{Per-skill accuracy shift of skill-based selection on \textbf{MATH-500}. Skill-based sampling improves weaker skills while preserving strong ones, flattening the accuracy curve toward balanced mastery. Skill-aware augmentation further enhances robustness across skills. (Detailed version is in Appendix \fref{fig:per_skill_acc_full})}
    \label{fig:per_skill_acc}
\vspace{-8pt}
\end{figure}

To examine the effect of our skill-based oversampling strategy on individual skills, we further evaluate 500 problems from the \textbf{MATH-500}~\citep{math500} benchmark. \fref{fig:per_skill_acc} reports the skill-wise accuracies across different settings (each position on the x-axis represents a distinct skill). Both random and skill-based sampling substantially improve accuracy over the base model. \textbf{Skill-based oversampling effectively aligns SFT data distribution with model weaknesses.} Weaker skills that are sampled more frequently correspond to larger accuracy gains.
Moreover, the accuracy of stronger skills remains high although sampled less frequently, suggesting that random sampling may waste training cost on areas where the model already performs well. Therefore, \textbf{skill-based training curve becomes notably flatter, showing that the model achieves more balanced and robust performance across skills}. Adding skill-aware augmentation further strengthens this effect, yielding even greater consistency in skill performance.

\subsection{Ablation Studies}
\label{sec:ablation}

\paragraph{Effect of Sampling Aggressiveness}
We examine how the aggressiveness of weakness sampling influences performance by varying the exponent of accuracy (replacing $\text{acc}^{-1}$ in the formula with $\text{acc}^{-T}$). As shown in Appendix Table~\ref{tab:ablation}, performance first rises quickly and then saturates as sampling becomes more aggressive. Thus, setting \(T=1.0\) provides a simple and effective balance.

\paragraph{Is the Full Chain of Skills Necessary?}
Our skill-aware SFT provides the full hierarchical skill chain. To test its necessity, we ablate the chain and expose only one layer, either top-level or leaf skills. As shown in Appendix Table~\ref{tab:ablation}, top-level skills yield little benefit, and leaf-level skills improve more but still underperform the full chain. This suggests that training with the complete skill tree structure is beneficial for model learning.

\subsection{Generalization Results}
Our approach also demonstrates strong generalization across a wider range of settings. Appendix~\ref{sec:generalization_extra} includes three complementary results: (i) generalization to alternative skill taxonomies and tree structures~\citep[EvalTree;][]{zeng2025evaltree}; (ii) generalization to different model families~\citep[R1-Distill-Llama-8B;][]{r1}; and (iii) gains over strong data-selection baselines, including LIMO~\citep{ye2025limo} and s1~\citep{s1}. These results support the broad applicability of our skill-aware distillation framework. \camerareadytext{These results provide further evidence that our Skill-Aware Data Selection and Fine-Tuning framework is broadly applicable and can serve as a general solution for data-efficient reasoning distillation.}

\section{Conclusion}

This research demonstrates that skill-based data selection and skill-aware training enable more capable, data-efficient, and interpretable reasoning distillation. By prioritizing examples from weaker skills of the student model and embedding explicit skill structures during SFT, our approach allows smaller models to acquire strong and robust reasoning abilities. These findings highlight the potential of skill-centric training as a general framework for improving distillation efficiency and transparency.

\section{Limitations}

While our study demonstrates clear benefits of skill-centric training, several limitations remain. First, our approach relies on existing skill trees. Although this structure covers most mathematical domains, it may not perfectly align with the skill decomposition used by the student model. Future work could explore more skill tree variants or automatically learned skill hierarchies. Second, the accuracy-based sampling assumes that per-skill evaluation reliably reflects model competence. However, skill-wise accuracy can be noisy, especially when each skill has limited evaluation data. A more robust estimate, perhaps through uncertainty modeling or multi-task validation, may improve stability in sampling decisions. Finally, we evaluate only two model scales (4B and 8B). Although results are consistent across three different models, further validation on larger sizes can be very helpful to assess generality.

\section{Ethical Considerations}

This work focuses on improving data efficiency and interpretability in reasoning model distillation and does not involve any human subjects or personally identifiable information. All datasets used, including OpenMathReasoning and benchmark test sets (e.g., AMC23, AIME, MATH, OlympiadBench), consist of publicly available mathematical problems without sensitive content. We emphasize that skill-centric training aims to enhance transparency and interpretability rather than automate human reasoning. Models trained with our framework should be deployed responsibly, with human oversight and clear communication of their limitations.

\section*{Acknowledgments}
This work is supported by computational resources and services provided by Advanced Research Computing (ARC), a division of Information and Technology Services (ITS) at the Unversity of Michigan, Ann Arbor.

\bibliography{custom}

\appendix

\section{Data Filtering Details}
\label{sec:data_filtering}

We use \textbf{OpenMathReasoning}~\citep{omr} as the primary dataset, a large math reasoning corpus containing 306K unique problems with 3.2M solutions sampled from DeepSeek-R1~\citep{r1}. From the corpus, we construct a 100K clean training pool by applying several filtering steps. First, we discard problems without a ground-truth answer. Each unique problem is associated with approximately ten candidate responses; we retain only those generated by DeepSeek-R1~\citep{r1} and only when the predicted final answer exactly matches the ground truth. For each problem, we then keep a single valid response to avoid duplication. This procedure yields roughly 105K problem–solution pairs. We ensure that there is no data leakage between our training corpus and the evaluation benchmarks. Finally, we randomly remove 5K instances to obtain a balanced set of 100K unique QA pairs used in our experiments.

\section{Skill Tree Attribution Details}
\label{sec:tree_attribution_details}
The prompt we used on \textit{Qwen/Qwen2.5-32B-Instruct}~\citep{qwen2.5} for top-down skill attribution are listed below:
{\small\begin{lstlisting}
Given the following Math problem:

Q&A: {qa_input}

Which of the following skills are involved to understanding or solving the problem? Even the most basic skills such as simple addition and subtraction must be taken into account. You can select multiple options if needed. Just return a list of skill names.

Skills:
{chr(10).join([f"- {name}" for name in child_names])}

Answer as a Python list of strings.
'''
\end{lstlisting}
}

\section{Generalization Results and Additional Comparisons}
\label{sec:generalization_extra}

This section reports additional experiments omitted from the main paper due to space limits. We study (1) sensitivity to the choice of the skill tree, (2) transfer to another model family, and (3) comparisons with existing data selection methods.

\subsection{Generalization to Different Skill Tree Design}
\label{sec:gen_skill_tree}
We evaluate whether our framework depends on the specific choice of skill tree. In the main paper we use Instruct-SkillMix~\citep{kaur2025instructskillmix}. Here we additionally test EvalTree~\citep{zeng2025evaltree}, whose hierarchy can be substantially deeper. Table~\ref{tab:evaltree_vs_ism} shows that EvalTree yields comparable or better results than Instruct-SkillMix across benchmarks, indicating that the method transfers across substantially different tree designs.

\begin{table*}[t]
\centering
\resizebox{\textwidth}{!}{%
\begin{tabular}{l l ccccc c}
\toprule
\textbf{Base Model} & \textbf{Data Selection} & \textbf{AMC23} & \textbf{AIME2024} & \textbf{AIME2025} & \textbf{MATH L5} & \textbf{OlympiadBench} & \textbf{Average} \\
\midrule
\multirow{4}{*}{\textbf{Qwen3-4B}}
& Base & 90.1 & 61.1 & 50.7 & 84.3 & 49.1 & 67.1 \\
& Random & 89.5 & 60.1 & 50.3 & 85.3 & 49.0 & 66.8 \\
& Instruct-SkillMix (1K) & 89.1 & \textbf{62.5} & 50.0 & \textbf{85.5} & \textbf{49.5} & 67.3 \\
& EvalTree (1K) & \textbf{90.3} & \textbf{62.5} & \textbf{53.3} & 85.4 & 49.4 & \textbf{68.2} \\
\midrule
\multirow{4}{*}{\textbf{Qwen3-8B}}
& Base & 88.2 & 61.1 & 50.2 & 84.7 & 49.1 & 66.7 \\
& Random & 90.2 & 62.6 & 50.8 & 86.0 & \textbf{50.7} & 68.1 \\
& Instruct-SkillMix (1K) & \textbf{93.4} & 62.1 & \textbf{51.3} & \textbf{86.2} & 49.7 & 68.5 \\
& EvalTree (1K) & 91.3 & \textbf{65.4} & \textbf{51.3} & 85.2 & 50.6 & \textbf{68.8} \\
\bottomrule
\end{tabular}}
\caption{Skill tree generalization using EvalTree~\citep{zeng2025evaltree} versus Instruct-SkillMix~\citep{kaur2025instructskillmix}.}
\label{tab:evaltree_vs_ism}
\end{table*}

\subsection{Generalization to Another Model Family}
\label{sec:gen_model_family}
To test generality beyond Qwen3, we run the same pipeline on \textit{deepseek-ai/DeepSeek-R1-Distill-Llama-8B}~\citep{r1}. Table~\ref{tab:llama_family} shows the same pattern as Qwen3: skill-based selection improves over random, and adding skill-aware SFT yields additional gains.

\begin{table*}[t]
    \centering
    \resizebox{\textwidth}{!}{%
        \begin{tabular}{l l l ccccc c}
        \toprule
        \textbf{Base Model} & \textbf{Data Selection} & \textbf{Fine-tuning Strategy} & \textbf{AMC23} & \textbf{AIME2024} & \textbf{AIME2025} & \textbf{MATH L5} & \textbf{OlympiadBench} & \textbf{Average} \\
        \midrule
        \multirow{6}{*}{R1-Distill-Llama-8B}
         & - & Base & 79.3 & \underline{37.1} & 27.8 & 61.5 & 38.5 & 48.8 \\
         & Full (100K) & Standard SFT & 74.8 & 30.9 & 21.7 & 59.0 & 36.2 & 44.5 \\
         & Random & Standard SFT & 81.0 & 36.1 & 30.4 & 70.3 & \textbf{42.4} & 52.0 \\
         & Random & Skill-aware SFT & \underline{82.0} & \textbf{37.8} & 30.9 & 71.2 & \underline{42.3} & \underline{52.8} \\
         & Skill-based & Standard SFT & 81.3 & 35.0 & \underline{31.0} & \textbf{71.9} & 42.2 & 52.3 \\
         & Skill-based & Skill-aware SFT & \textbf{83.1} & 36.3 & \textbf{31.3} & \underline{71.8} & \underline{42.3} & \textbf{53.0} \\
        \bottomrule
        \end{tabular}
    }
\caption{Accuracy (\%) of DeepSeek-R1-Distill-Llama-8B under different training data selection and fine-tuning strategies using \textbf{1K training examples}. Each column is \textbf{bolded} at its highest value and \underline{underlined} at its second highest. Results are reported using Avg@8 across five math benchmarks.}
\label{tab:llama_family}
\end{table*}

\subsection{Comparison with Existing Data Selection Methods}
\label{sec:compare_existing}
While improvements can be numerically small, improving strong reasoning models with only 1,000 SFT examples is inherently challenging. To contextualize this setting, Table~\ref{tab:compare_limo_s1} compares our method with LIMO~\citep{ye2025limo} and s1~\citep{s1} on Qwen3-4B. We evaluate LIMO using its released dataset (offline), since code is not released. For s1, we include both the released dataset (offline) and rerunning their selection code on our 100K pool (online). Existing selections degrade performance or match random sampling, whereas our method consistently improves across benchmarks.

\begin{table}[t]
\centering
\resizebox{\columnwidth}{!}{%
\begin{tabular}{lccccc}
\toprule
\textbf{Method} & \textbf{AMC23} & \textbf{AIME2024} & \textbf{AIME2025} & \textbf{MATH L5} & \textbf{Average} \\
\midrule
Base & 90.1 & 61.1 & 50.7 & 84.3 & 71.6 \\
Random & 89.5 & 60.1 & 50.3 & 85.3 & 71.3 \\
LIMO (offline) & 88.4 & \textbf{61.3} & 45.4 & 85.3 & 70.1 \\
s1 (offline) & 89.1 & 57.9 & 47.5 & \textbf{85.9} & 70.1 \\
s1 (online) & 89.1 & 60.7 & 50.2 & 84.9 & 71.2 \\
Our method & \textbf{91.9} & \textbf{64.6} & \textbf{50.8} & 85.3 & \textbf{73.2} \\
\bottomrule
\end{tabular}}
\caption{Comparison with existing data selection methods on Qwen3-4B under our 1K SFT setting. Offline uses released datasets, online reruns the selection code on our 100K pool.}
\label{tab:compare_limo_s1}
\end{table}

\section{Compute Cost}
\label{sec:compute_cost}

We report the compute cost of the two main stages in our pipeline: (i) one-time skill labeling (Skill Tree Attribution) over the 100K training pool, and (ii) SFT on the selected 1K subset.

\paragraph{One-time skill labeling (inference-only).}
Skill attribution is performed by prompting Qwen2.5-32B-Instruct to assign each example to a skill chain via top-down traversal (Appendix~\ref{sec:tree_attribution_details}). This stage is \emph{inference-only} and is \emph{model-agnostic} with respect to the student: the labels depend only on the skill definitions and the data, not on the target model being fine-tuned. In our implementation, labeling the 100K pool took approximately \textbf{200 GPU hours} in total. Importantly, this labeling cost is \emph{amortizable}: once produced, the same labels can be reused across multiple student models, seeds, and future experiments under the same skill taxonomy. We note that a comparable corpus-level inference cost is common in recent data selection work, which often relies on LLMs to generate auxiliary metadata (e.g., quality, difficulty, topic, or other annotations) for a large candidate pool before selecting a small training subset~\citep{ye2025limo, s1, li2025mass}.

\paragraph{SFT cost.}
For SFT, our default setting fine-tunes each student model for 5 epochs (Appendix~\ref{sec:training-details}). Under our 1K-example setting, each training run costs about \textbf{40 GPU hours} per run. For the \textbf{full 100K} standard SFT baseline, the training cost is substantially higher; in our setup it is approximately \textbf{512 GPU hours}. This comparison highlights that, even when accounting for the one-time labeling cost, our pipeline remains compute-efficient in repeated use cases (e.g., evaluating multiple students or running multiple ablations), because the labeling step does not scale with the number of students.

\section{Training Details}
\label{sec:training-details}

\paragraph{Environment.}
All experiments were conducted using NVIDIA A40
GPUs with 48GB memory. The software environment was configured as follows:
\begin{itemize}
  \item \texttt{360-LLaMA-Factory}~\citep{360-llama-factory} (A long-CoT adapted version of \texttt{LLaMA-Factory}  0.9.1~\citep{llamafactory})
  \item \texttt{torch} 2.7.0
  \item \texttt{transformers} 4.51.3
  \item \texttt{accelerate} 1.0.1
  \item \texttt{datasets} 3.1.0
  \item \texttt{trl} 0.9.6
  \item \texttt{peft} 0.12.0
  \item \texttt{deepspeed} 0.14.4
\end{itemize}

\paragraph{SFT Training.}
For SFT training, we used the following settings:
\begin{itemize}
  \item Batch size: 32 (8 GPUs * 4 Gradient Accumulation)
  \item Epoch: 5
  \item Learning rate: 1e-5
  \item Optimizer: AdamW
  \item Learning rate scheduler: cosine with warmup
  \item Warmup ratio: 0.1
  \item Cutoff length: 8192
  \item Time Cost: 4 hours per run
\end{itemize}

\paragraph{Decoding Setup.}
During inference, we applied the following decoding settings:
\begin{itemize}
  \item Temperature: 0.6
  \item Max tokens: 16384
  \item Top-p: 0.95
\end{itemize}

\section{Additional Experiment Results}

\begin{figure*}[tb!]
    \centering
    \includegraphics[width=\textwidth]{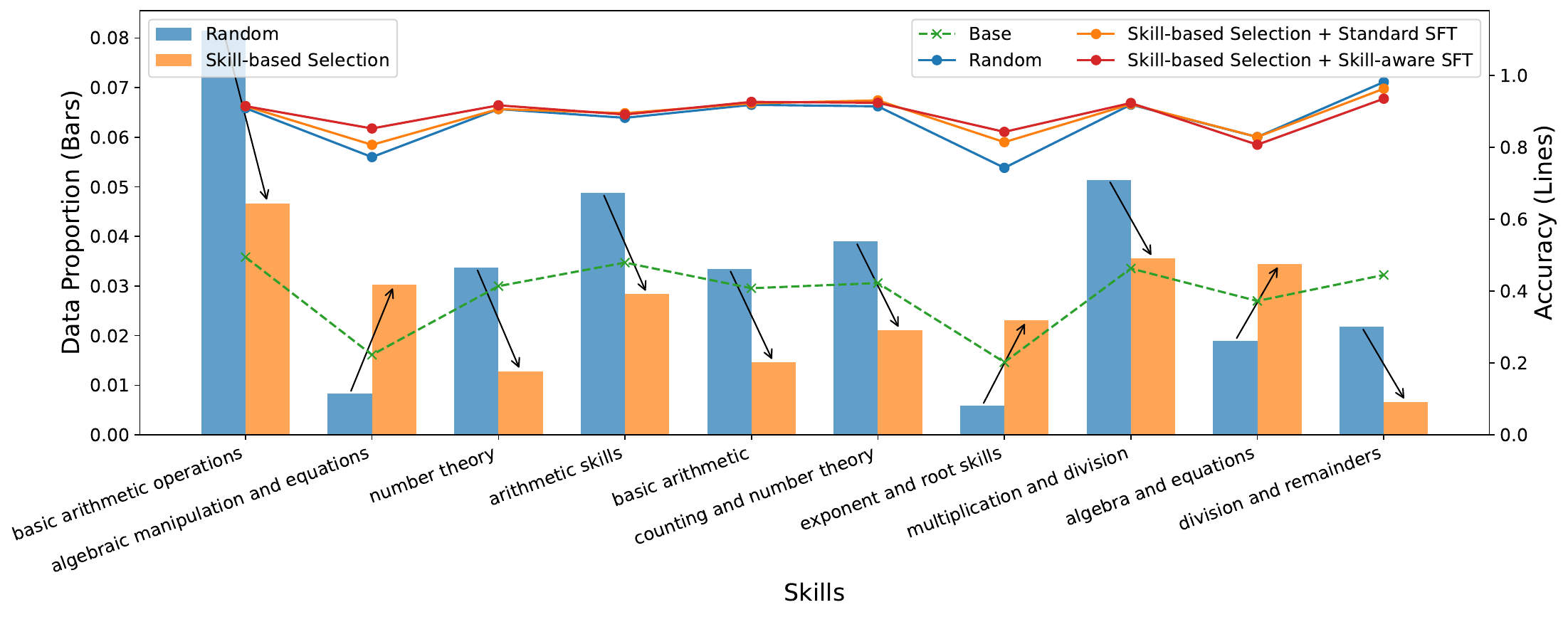}
        \caption{Data proportion shift of skill-based selection and per-skill accuracy (\%) on \textbf{MATH-500}. Skill-based sampling improves weaker skills while preserving strong ones, flattening the accuracy curve toward balanced mastery. Skill-aware augmentation further enhances robustness across skills.}
    \label{fig:per_skill_acc_full}
\end{figure*}

Detailed version of \fref{fig:per_skill_acc} with data proportion shift is shown in \fref{fig:per_skill_acc_full}.

Simple version of the Ablation Study in Section~\ref{sec:ablation} is shown in \tref{tab:ablation}, and its full version is shown in \tref{tab:ablation_full}.

\begin{table}[t]
\centering
\footnotesize
\setlength{\tabcolsep}{6pt}
\begin{tabular}{cc}
\toprule
\textbf{Setting} & \textbf{Avg Accuracy} \\
\midrule
\multicolumn{2}{l}{\textit{Effect of Sampling Aggressiveness}} \\
\quad $T=0.5$ & 70.7 \\
\quad $T=0.75$ & 71.3 \\
\quad $\mathbf{T=1.0}$ & 71.9 \\
\quad $T=2.0$ & 72.0 \\
\quad $T=3.0$ & 71.9 \\
\midrule
\multicolumn{2}{l}{\textit{Is the Full Skill Chain Necessary?}} \\
\quad \textbf{Full skill chain} & 72.9 \\
\quad Root Skills Only & 72.2 \\
\quad Leaf Skills Only & 72.7 \\
\bottomrule
\end{tabular}
\caption{Ablations on Sampling Aggressiveness and Hierarchical Skill Chain. Default settings are \textbf{bolded}. Full results are in Appendix \tref{tab:ablation_full}.}
\label{tab:ablation}
\vspace{-6pt}
\end{table}

\begin{table*}[t]
\centering
\resizebox{\textwidth}{!}{
\begin{tabular}{l l ccccc}
\toprule
\textbf{Ablation} & \textbf{Setting} & \textbf{AMC23} & \textbf{AIME2024} & \textbf{AIME2025} & \textbf{MATH L5} & \textbf{Average} \\
\midrule
\multirow{4}{*}{Effect of Sampling Aggressiveness}
 & $T=0.5$ & 89.7 & 60.0 & 47.9 & 85.2 & 70.7 \\
 & $T=1.0$ & 89.1 & \textbf{62.5} & \textbf{50.0} & \underline{85.7} & \underline{71.9} \\
 & $T=2.0$ & \underline{90.6} & \textbf{62.5} & \underline{48.8} & \textbf{85.9} & \textbf{72.0} \\
 & $T=3.0$ & \textbf{91.6} & \underline{61.7} & \underline{48.8} & 85.6 & \underline{71.9} \\
\midrule
\multirow{3}{*}{Is the Full Skill Chain Necessary?}
 & Full skill chain & \textbf{91.9} & \textbf{64.2} & \underline{50.4} & 85.1 & \textbf{72.9} \\
 & Root Skills Only & \underline{91.6} & 58.3 & \textbf{52.5} & \underline{86.3} & 72.2 \\
 & Leaf Skills Only & 90.9 & \underline{62.9} & 50.0 & \textbf{86.9} & \underline{72.7} \\
\bottomrule
\end{tabular}}
\caption{Ablations on sampling aggressiveness ($T$) and on exposing different portions of the skill hierarchy during skill-aware SFT. Within each ablation block, the highest value per column is \textbf{bolded} and the second-highest is \underline{underlined}.}
\label{tab:ablation_full}
\end{table*}

\section{Use of Large Language Models}
We acknowledge that we only used LLMs to check grammatical errors in the paper and to improve the clarity of expression.

\end{document}